\documentclass[letterpaper, 10 pt, conference]{ieeeconf}  

\IEEEoverridecommandlockouts                              

\usepackage{xcolor}
\usepackage{graphicx}
\usepackage{biblatex}
\usepackage{url}
\usepackage[hidelinks]{hyperref}
\usepackage{amsmath,amssymb,amsfonts}
\title{\LARGE \bf
Robust Mitosis Detection Using a Cascade Mask-RCNN Approach With Domain-Specific Residual Cycle-GAN Data Augmentation}

\author{Gauthier Roy$^{1*}$, Jules Dedieu$^{1*}$, Capucine Bertrand$^{1}$, Alireza Moshayedi$^{1}$, Ali Mammadov$^{1}$, St\'ephanie Petit$^{1}$, \\Saima Ben Hadj$^{1}$ and Rutger H.J. Fick$^{1}$\\ \\ $^{1}$Tribvn Healthcare, Paris, France

\thanks{$^{*}$These Authors contributed equally to the work.}%
}

\begin{document}

\maketitle
\thispagestyle{empty}
\pagestyle{empty}

\begin{abstract}
For the MIDOG mitosis detection challenge, we created a cascade algorithm consisting of a Mask-RCNN detector, followed by a classification ensemble consisting of ResNet50 and DenseNet201 to refine detected mitotic candidates. The MIDOG training data consists of 200 frames originating from four scanners, three of which are annotated for mitotic instances with centroid annotations. Our main algorithmic choices are as follows: first, to enhance the generalizability of our detector and classification networks, we use a state-of-the-art residual Cycle-GAN to transform each scanner domain to every other scanner domain. During training, we then randomly load, for each image, one of the four domains. In this way, our networks can learn from the fourth non-annotated scanner domain even if we don't have annotations for it. Second, for training the detector network, rather than using centroid-based fixed-size bounding boxes, we create mitosis-specific bounding boxes. We do this by manually annotating a small selection of mitoses, training a Mask-RCNN on this small dataset, and applying it to the rest of the data to obtain full annotations. We trained the follow-up classification ensemble using only the challenge-provided positive and hard-negative examples. On the preliminary test set, the algorithm scores an F1 score of 0.7578, putting us as the second-place team on the leaderboard.
\end{abstract}

\section*{Introduction}
Mitosis detection is a highly challenging task in pathology due to the rarity of the events and the highly variable morphological appearance of a cell undergoing mitosis - some being very clear and others highly ambiguous. While several mitosis detection challenges have been organized over the past years (MITOS12, MITOS14, AMIDA, TUPAC), none of them focused on testing the effect of domain shift on the robustness of a mitosis detection method. The MIDOG challenge~\cite{midog} specifically addresses this by providing training data originating from four different scanners but making the unseen test set (partially) consist of images that are not from these same scanners.

\subsection*{Dataset}
Following the challenge description: the MIDOG training subset consists of 200 whole slide images (WSI) from human breast cancer tissue samples stained with routine H\&E dye. The samples were digitized with four slide scanning systems: the Hamamatsu XR NanoZoomer 2.0, the Hamamatsu S360, the Aperio ScanScope CS2 and the Leica GT450, resulting in 50 WSI per scanner. For the slides of three scanners, a selected field of interest sized approximately 2mm$^2$ (equivalent to ten high power fields) was annotated for mitotic figures and hard negative look-alikes. These annotations were collected in a multi-expert blinded set-up. For the Leica GT450, no annotations were available. The preliminary test set consists of five WSI each for four undisclosed slide scanning systems of which only two were also part of the training set. This preliminary test set was used for evaluating the algorithms prior to submission and publishing preliminary results on a leaderboard. The final test set consists of 20 additional WSI from the same four scanners used for the preliminary test set. The evaluation through a Docker-based submission system ensured that the participants had no access to the (preliminary) test images during method development.

\section*{Material and Methods}
We base our method around a classic cascade approach to detect mitotic instances in H\&E-stained images. We first use a Mask-RCNN~\cite{he2017mask} to detect mitotic candidates in an image. These candidates are then extracted as small patches and given to a classifier ensemble of a ResNet50~\cite{he2016deep} and DenseNet201~\cite{huang2017densely}. The predictions are merged via weighted average and the final score is returned.

\begin{figure}
    \centering
    \includegraphics[width=0.5\textwidth]{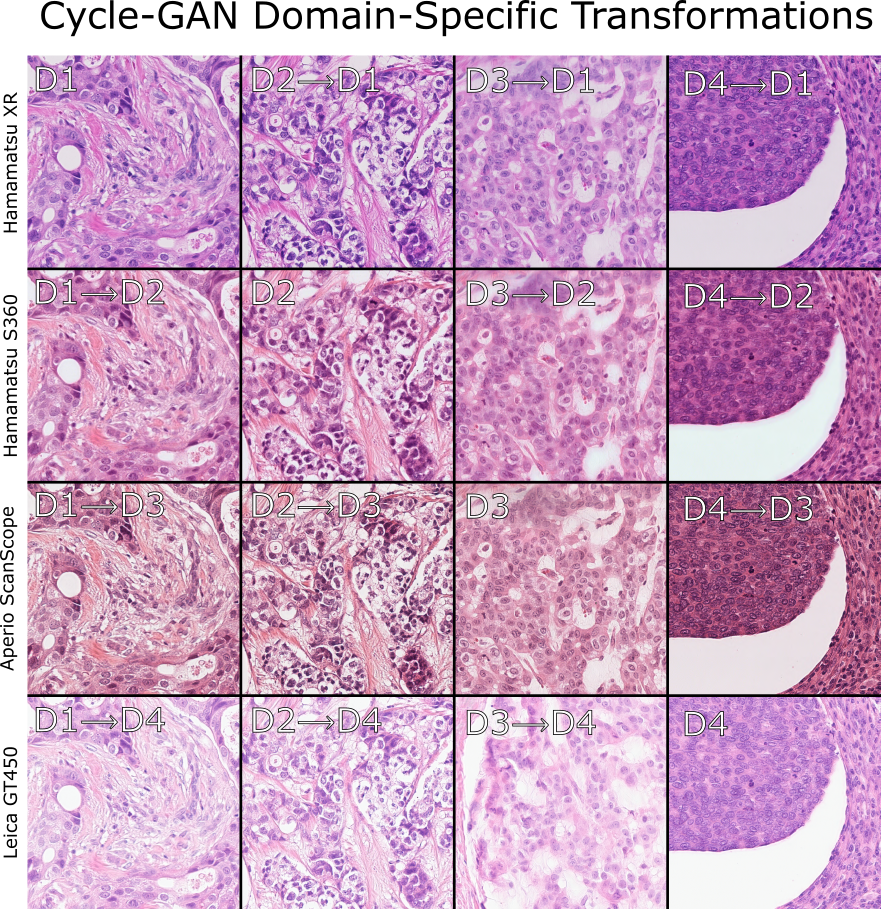}
    \caption{Residual Cycle-GAN transformed patches. The diagonal are original patches, off-diagonal patches are domain-transformed.}
    \label{fig:GAN}
\end{figure}

To improve the generalizability of the method - which is the main purpose of the challenge - we used a Residual Cycle-GAN~\cite{de2021residual} to transform each image of the training images into all other available domains. In this way, each mitotic annotation is available in all 4 scanner domains. This differs from standard data augmentation (color, hue, brightness, etc.), in that these are not random shifts in appearance for the training process, but specifically towards domains that we \emph{know} are in the testing set. In Figure \ref{fig:GAN} we show a 4x4 grid of images of the 4 domains that we transformed to all other domains.

To improve the information present in the data for training a detector, we use Mask-RCNN to create pixel-wise annotations for all annotated mitotic instances. Since we know where all mitoses are, we use Inkscape to manually annotate the first 100 or so, train a pretrained Mask-RCNN model on this small dataset, and apply it specifically around other known mitoses. We use test-time augmentation (8 rotations and flips) and average the predicted masks for each mitosis, resulting in clean masks for most annotations. The remaining "difficult" cases were manually completed, providing us mitosis-specific bounding boxes for all mitotic instances. The average bounding box diagonal in the dataset is 28.8$\pm$7.9 pixels, which is consistent with the MITOS12 dataset~\cite{kausar2020smallmitosis}.

\subsection*{Domain-Specific Residual Cycle-Gan Augmentation}
For the Residual Cycle-GAN~\cite{de2021residual} we followed the reference implementation of two sets of generators and discriminators. The residual Cycle-GAN follows the same principle as a regular one, with the difference that the input image has a direct skip connection with the generated output image. In this way, the generator does not need to reconstruct the image from a set of filter outputs, but only needs to add a "residual", i.e. a color change in the input image so that it resembles a target domain. As it reduces the computational load on the generator, this approach requires fewer data and converges more quickly, see \cite{de2021residual} for more details.

\subsection*{Network Training}
We split our training data into 45 training slides and 5 validation slides per scanner, ensuring that the validation set had both highly mitotic and non-mitotic slides. The Torchvision implementation of Mask-RCNN was pretrained on the public COCO2017 dataset, and both the ResNet50 and DenseNet201 were pretrained on ImageNet. For Mask-RCNN training, we used a patch size of 3000$\times$3000 pixels and a batch size of 1. We did not train on patches that did not contain any mitoses. We found that using a larger patch size improves the validation performance, and did not improve when adding negative patches. We augmented Mask-RCNN training using skewing, 8 random flips/mirroring, and the domain-specific Cycle-Gan augmentation stated before. We used SGD with a plateau-reduction learning rate scheduler starting at 0.002 and reducing by a factor of 2 if the PR-AUC does not improve after 5 epochs. We warmed up the optimizer during the first epoch and only unfroze the last two convolutional blocks of the Mask-RCNN network. We ran the algorithm until convergence after around 200 epochs.

The classification networks were only trained with the positive and negative instances provided by the challenge organizers - we found that adding hard negatives detected by the detector did not improve leaderboard performance. We used a batch size of 32, and trained for 100 epochs, and kept the model with the best F1 score. We used ADAM with standard parameters and a Cosine annealing learning rate scheduler starting at $2\times10^{-5}$ with a focal loss. For both networks, we only unfroze the backbone after 5 epochs. We used a patch size of 80$\times$80, which we resized to 224$\times$224 to conform with ImageNet pretraining. We used our GAN-based domain augmentation, together with H\&E specific data augmentation \cite{faryna2021tailoring}, with parameters $n=3$, $m=7$. The classification head consists of 3 blocks of convolutions with Relu, batch normalization, and dropout set to 0.5, followed by a fully connected layer to the output.

\section*{Evaluation and Results}
For both the detector and classifiers, many variations of optimization parameters were tried and the model with the best PR-AUC on validation was selected. On validation, our PR-AUC was 0.8823 and F1 was 0.8287. On the preliminary test set, we scored an F1-score of 0.7577, resulting from a 0.7820 precision and a 0.7349 recall. For reference, this put us at the second-highest scoring team.

\printbibliography

\end{document}